\documentclass[11pt]{article}

\usepackage[preprint]{acl}

\usepackage{times}
\usepackage{latexsym}

\usepackage[T1]{fontenc}

\usepackage[utf8]{inputenc}

\usepackage{microtype}

\usepackage{inconsolata}

\usepackage{graphicx}

\usepackage{booktabs}
\usepackage{multirow}
\usepackage{enumitem}
\usepackage{algorithm}
\usepackage{algpseudocode}
\usepackage{amsmath} 
\usepackage{subcaption}
\usepackage{xcolor}
\usepackage[table]{xcolor}
\newcommand{\framework}{\textsc{Elmes}\textsuperscript{+}}

\newcommand{\scenegen}{\textsc{SceneGen}}

\newcommand{\eg}{\textit{e.g.}}

\title{\framework{}: Automated Construction of Fine-Grained Evaluation Rubrics \\for Large Language Models in Long-Tail Educational Scenarios}

\author{
  Tao Liu\textsuperscript{1,2},
  Ye Lu\textsuperscript{1,2},
  Ruohua Zhang\textsuperscript{1,2},
  Siyu Song\textsuperscript{1,2},
  Wentao Liu\textsuperscript{1,2,3},
  Aimin Zhou\textsuperscript{1,2,3},
  and Hao Hao\textsuperscript{1,*} \\
  \textsuperscript{1}Shanghai Institute of AI for Education, East China Normal University, Shanghai 200062, China \\
  \textsuperscript{2}School of Computer Science and Technology, East China Normal University, Shanghai 200062, China \\
  \textsuperscript{3}Shanghai Innovation Institute, Shanghai 200231, China \\
  {\small \textbf{Corresponding author:} Hao Hao (\texttt{hhao@mail.ecnu.edu.cn})}
}

\begin{document}
\maketitle

\begin{abstract}
Evaluating large language models (LLMs) for education requires measuring how models teach, not only what they know. Existing benchmarks emphasize domain-general correctness or depend on manually designed rubrics that scale poorly to long-tail pedagogical scenarios. We introduce \framework{}, an end-to-end framework for constructing, refining, and applying fine-grained scenario-specific rubrics. \framework{} combines a declarative multi-agent engine for teacher--student--judge interactions with \scenegen{}, a self-evolving module that co-optimizes evaluation criteria and test data from expert-defined pedagogical dimensions. Using \framework{}, we build Edu-330, covering 330 scenarios across 11 subjects, 3 grade bands, and 10 task types, with over 1{,}000 second-level indicators. Experiments on Edu-330 and four expert-authored gold-standard scenarios show that educational capability is multidimensional: top-tier LLMs differ mainly in creativity and values integration, knowledge-strong models may fail at Socratic scaffolding, and the education-specialized InnoSpark achieves the best human-evaluated average score. LLM judges preserve human-comparable rankings with much lower scoring variance, but exhibit judge-specific biases such as self-preference. Ablations show that expert-scored few-shot anchoring improves human--LLM alignment, while reasoning enforcement and greedy decoding are model-dependent. \framework{} thus provides scalable diagnostic infrastructure for pedagogically grounded LLM evaluation.
\end{abstract}

\section{Introduction}
\label{sec:intro}
Large language models have rapidly entered educational technology, powering intelligent tutoring, lesson planning, and personalized content generation~\citep{kasneci2023chatgpt, yan2024practical}.
Yet evaluating whether an LLM is genuinely effective as a pedagogical agent remains difficult: unlike factual question answering, educational tasks are subjective, multi-dimensional, and context-dependent, requiring models to balance content accuracy, scaffolding, emotional sensitivity, and cultural appropriateness~\citep{wang2024large}.

Existing evaluation paradigms remain insufficient.
General-purpose benchmarks such as MMLU-Pro~\citep{wang2024mmlu}, C-Eval~\citep{huang2023c}, and GSM8K~\citep{cobbe2021training} emphasize knowledge or reasoning rather than pedagogical process quality; interactive frameworks such as MT-Bench~\citep{zheng2023judging} and Chatbot Arena~\citep{chiang2024chatbot} provide coarse preference signals; and education-specific benchmarks~\citep{dan2023educhat, xu2025edubench} cover limited scenarios with manually curated rubrics.

The core challenge is the rubric construction bottleneck: fine-grained criteria require pedagogical expertise, while educational scenarios grow combinatorially with subject, grade, and task type.

We address this challenge with \framework{}, which automates educational LLM evaluation from rubric generation to scoring and refinement.
Our design principle is that evaluation metrics and test data should co-evolve: score distributions can reveal overly strict, overly lenient, or weakly discriminative rubrics, while test data can be regenerated to better exercise target pedagogical dimensions.

\framework{} makes the following contributions:

\begin{enumerate}[leftmargin=*]
\item \textbf{Multi-Agent Evaluation Engine.}
We design a declarative YAML-to-DAG engine that supports teacher--student--judge role configurations, conditional routing, trajectory-level scoring, and expert-in-the-loop scenario refinement for both single-turn and multi-turn educational tasks.

\item \textbf{Self-Evolving Rubric Synthesis (\scenegen{}).}
We introduce a closed-loop synthesis module that starts from 4 expert-defined dimensions and co-evolves rubrics and test data by diagnosing overly strict, overly lenient, or low-discrimination metrics, producing over 1{,}298 second-level indicators for 330 scenarios.

\item \textbf{Large-scale Benchmark and Expert-level Evaluation.} 
We construct \textbf{Edu-330}, a benchmark spanning 330 core scenarios, and complement it with 4 expert-designed pedagogical scenarios to reveal model-specific educational profiles and trade-offs overlooked by domain-agnostic benchmarks.

\item \textbf{Consistency Enhancement and Human-AI Alignment.} 
We analyze and improve LLM-as-a-judge reliability through expert anchoring and reasoning constraints, reducing mean-score deviation from human experts by approximately 30\% while maintaining high inter-rater reliability.
\end{enumerate}

\section{Related Work}
\label{sec:related}

\subsection{General-Purpose LLM Benchmarks}

LLM evaluation has been dominated by knowledge- and reasoning-centric benchmarks.
MMLU~\citep{hendrycks2021measuring} and MMLU-Pro~\citep{wang2024mmlu} assess broad factual knowledge, C-Eval~\citep{huang2023c} targets Chinese-language knowledge understanding, and GSM8K~\citep{cobbe2021training} and MATH~\citep{hendrycks2021measuring} evaluate mathematical reasoning.
While effective at measuring what a model knows, these benchmarks do not assess how well it can teach, which requires evaluating pedagogical process quality rather than outcome correctness alone.

\subsection{Interactive and Judge-Based Evaluation}

The LLM-as-a-Judge paradigm~\citep{zheng2023judging} addresses the scalability limitations of human evaluation by using strong LLMs to assess weaker ones.
MT-Bench~\citep{zheng2023judging} evaluates multi-turn conversational quality, while Chatbot Arena~\citep{chiang2024chatbot} crowdsources pairwise preferences at scale.
However, their holistic or pairwise preference signals lack the dimensional granularity needed for educational diagnostics; a model may ``win'' overall while failing in specific pedagogical dimensions (\eg, metacognition cultivation or emotional support).

Several studies have documented systematic biases in LLM judges, including position bias, verbosity bias, and self-preference~\citep{wang2026large, zheng2023judging}.
Prometheus~\citep{kim2024prometheus} trains dedicated judge models with fine-grained, rubric-based evaluation.
Auto-J~\citep{li2024generative} generates critique-based training data for judge models.
We build on this line of work through multi-judge ensembling and expert-anchored calibration, while automatically generating rubrics rather than assuming them as given.

\subsection{Education-Specific LLM Evaluation}

Recent work has begun addressing educational evaluation specifically.
EduChat~\citep{dan2023educhat} provides a pedagogically aligned chatbot but lacks systematic evaluation rubrics, while EduBench~\citep{xu2025edubench} focuses on educational knowledge rather than teaching process quality.
EQGBench~\citep{zhou2025answers} targets educational question generation, MathDial~\citep{macina2023mathdial} evaluates mathematical tutoring dialogues, and TutorBench~\citep{srinivasa2025tutorbench} assesses tutoring quality with process-level metrics but fixed, manually designed rubrics.

These efforts still cover limited educational scenarios and often depend on manually designed rubrics, making extension difficult.
\framework{} addresses both limitations through automated rubric generation and iterative refinement that can scale to hundreds of scenarios with minimal expert effort.

\section{The \framework{} Framework}
\label{sec:method}
\begin{figure*}[ht!]
\centering
\includegraphics[width=0.9\textwidth]{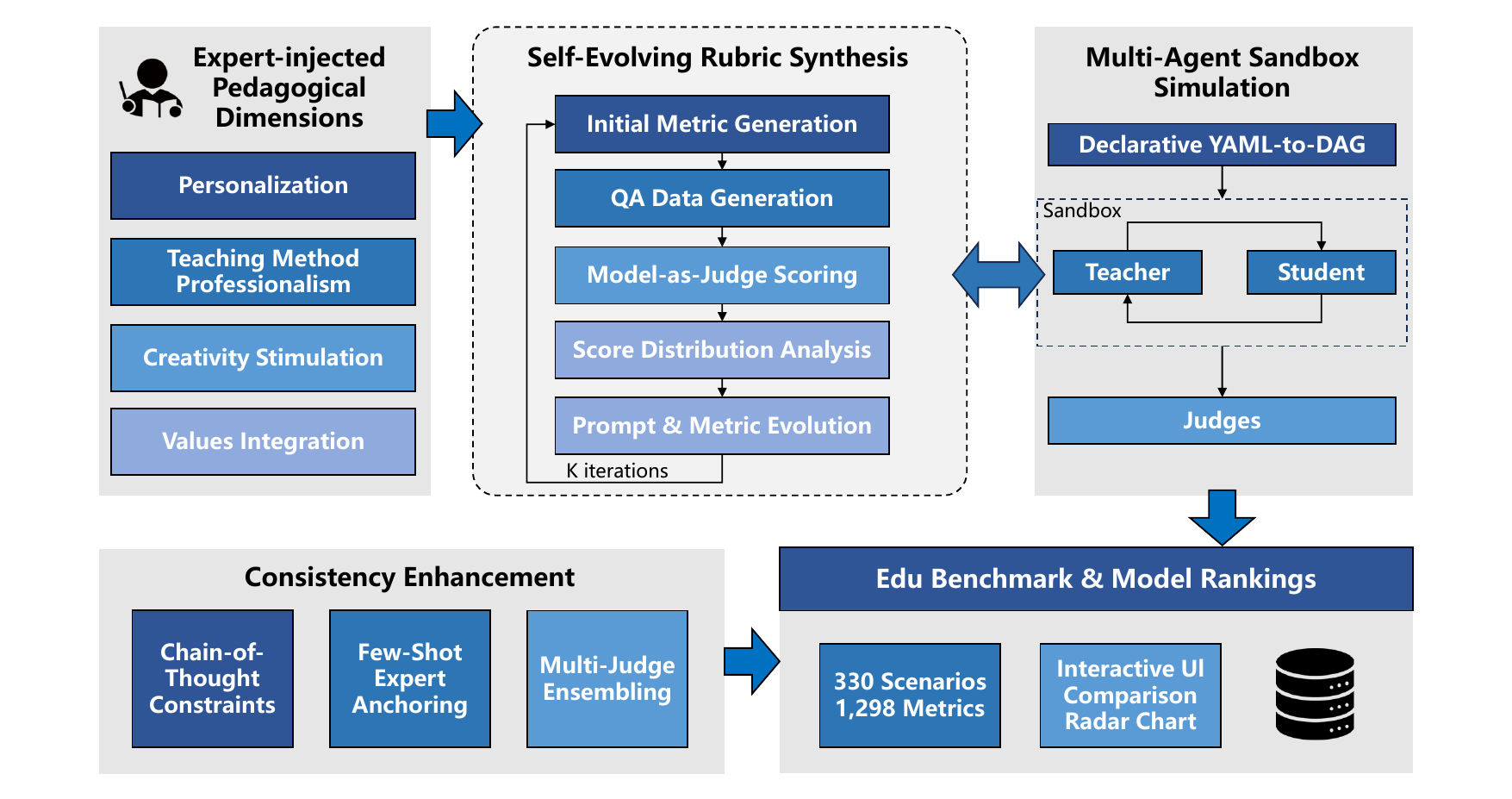}
\caption{Overview of \framework{}. \scenegen{} evolves rubrics and test data from expert-defined pedagogical dimensions, while the multi-agent engine orchestrates interactions among teachers, students, and judges and improves consistency with expert anchoring and multi-judge ensembling.}
\label{fig:elmes-framework}
\end{figure*}

\framework{} comprises two integrated components (Figure~\ref{fig:elmes-framework}): a multi-agent evaluation engine~(\S\ref{sec:engine}) and a self-evolving rubric synthesis module~(\S\ref{sec:scenegen}).

\subsection{Multi-Agent Evaluation Engine}
\label{sec:engine}

\subsubsection{Declarative Configuration System}

A central design principle of \framework{} is to specify evaluation scenarios declaratively rather than programmatically.
Each YAML file defines an educational scenario by binding a model to an instructional role, injecting task-specific content, specifying the interaction flow, and attaching a rubric-based judge, allowing experts to adapt scenarios without changing framework code.
Figure~\ref{fig:yaml-scenario} shows a simplified question-generation scenario.

\begin{figure}[ht!]
\centering
\setlength{\fboxsep}{8pt}
\fcolorbox{gray!35}{gray!8}{
\begin{minipage}{0.94\linewidth}
\scriptsize\ttfamily
models:\\
\hspace*{1em}test\_model: \\
\hspace*{1em}judge\_model: \\[2pt]
agents:\\
\hspace*{1em}test\_model:\\
\hspace*{2em}model: test\_model\\
\hspace*{2em}role: educational question generator\\
\hspace*{2em}{\color{blue}\textbf{content: "Generate a contextualized math}}\\
\hspace*{2em}{\color{blue}\textbf{problem for \{question\}"}}\\[2pt]
tasks:\\
\hspace*{1em}mode: iter\\
\hspace*{1em}content:\\
\hspace*{2em}{\color{blue}\textbf{- question: "Design a Grade 4 problem on}}\\
\hspace*{2em}{\color{blue}\textbf{equilateral-triangle perimeter ...."}}\\
\hspace*{2em}{\color{red}\textbf{(Generated by \scenegen{})}}\\[2pt]
directions: [START to test\_model, test\_model to END]\\[2pt]
evaluation:\\
\hspace*{1em}model: judge\_model\\
\hspace*{1em}format: [{\color{red}\textbf{ Generated by \scenegen{}}}]
\end{minipage}}
\caption{Simplified YAML configuration for defining an educational question-generation scenario.}
\label{fig:yaml-scenario}
\end{figure}

\subsubsection{DAG-Based Interaction and Evaluation}

The \texttt{directions} specification is compiled into a LangGraph state graph, where each node represents an agent and edges define the conversation flow.
Edges can either forward one agent's output directly to the next agent or invoke a router that selects the next node from the current dialogue state, with built-in support for keyword-based termination and multi-condition gates.
For multi-turn scenarios such as Socratic tutoring, teacher and student agents interact in a loop controlled by a maximum turn budget $T_{\max}$ and task-specific exit conditions.
The resulting dialogue trajectory $\tau = (m_1, m_2, \ldots, m_n)$ is then evaluated by the judge agent under the scenario-specific rubric.
Given a rubric $\mathcal{R} = \{(d_i, w_i)\}_{i=1}^{D}$, where each $d_i$ is a pedagogical evaluation dimension generated or refined by \scenegen{} and $w_i$ is its weight, the overall score is computed as:
\begin{equation}
\text{Score}(\tau, \mathcal{R}) = \sum_{i=1}^{D} w_i \cdot s_i(\tau, d_i)
\label{eq:score}
\end{equation}
where $s_i(\tau, d_i)$ denotes the judge-assigned score for trajectory $\tau$ on dimension $d_i$.

\subsubsection{Interactive Human Expert Interface for Scenario Configuration}
\label{sec:expert-interface}

To support expert-in-the-loop refinement, \framework{} provides a Gradio-based interface that lets educators and curriculum designers load, inspect, edit, and rerun scenario configurations without writing code.
Experts can inject domain knowledge into \scenegen{}-generated rubrics, adjust vague or misaligned criteria, and calibrate test cases through immediate evaluation feedback.
In practice, educators typically converge after 2--3 rounds of configuration updates.
The full interface is shown in Appendix~\ref{sec:appendix-ui}, Figure~\ref{fig:appendix-ui}.

\subsection{Self-Evolving Rubric Synthesis: \scenegen{}}
\label{sec:scenegen}
Another key innovation of \framework{} is \scenegen{}, a module that automatically generates and iteratively refines evaluation rubrics for educational scenarios. The core design is to let evaluation metrics and test data co-evolve: deficiencies in one can be detected through analysis of the other and corrected in subsequent iterations.

\begin{figure*}[ht!]
\centering
\includegraphics[width=0.9\textwidth]{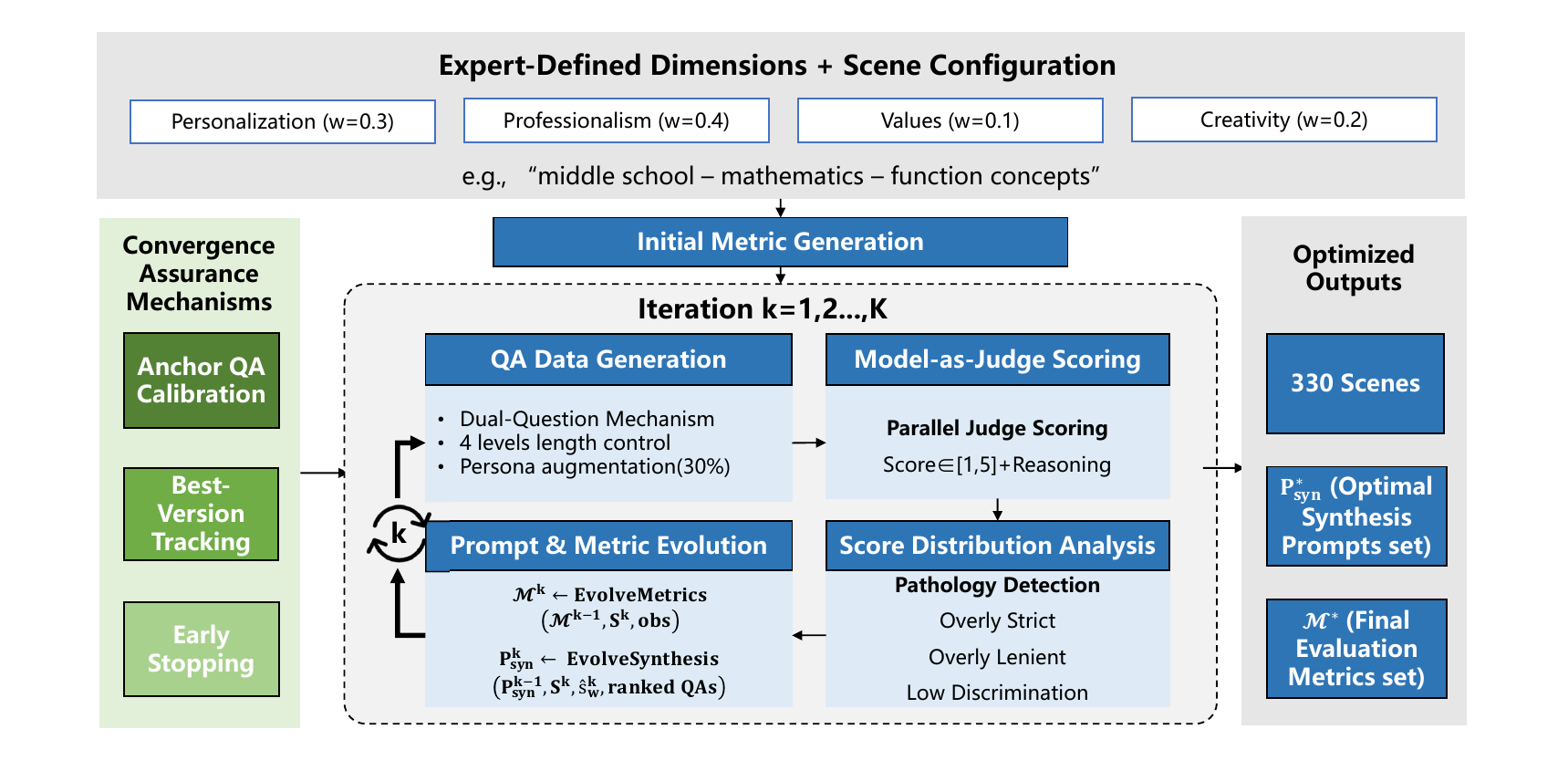}
\caption{\scenegen{} workflow. From expert-defined dimensions, this pipeline generates initial metrics and iteratively co-refines synthesis prompts and evaluation criteria with anchor calibration, early stopping, and best-version rollback.}
\label{fig:scenegen}
\end{figure*}

\subsubsection{Expert-Defined Dimensional Framework}

Rather than generating evaluation criteria from scratch, \scenegen{} begins with a compact set of initial pedagogical dimensions.
In this work, the initial dimensions are derived from no fewer than 70 interviews with education experts and grounded in established theories, including Shulman's pedagogical content knowledge (PCK), Vygotsky's zone of proximal development, and the TPACK model~\citep{vygotsky1978mind,mishra2006technological}.
This process yields four starting dimensions with expert-assigned weights:

\begin{itemize}[leftmargin=*, itemsep=1pt, topsep=2pt, parsep=0pt]
\item \textbf{Personalization} ($w_1 = 0.3$): Adaptation to learner profiles, cognitive levels, and individual needs.
\item \textbf{Teaching Method Professionalism} ($w_2 = 0.4$): Quality of pedagogical strategies, scaffolding, and domain expertise.
\item \textbf{Creativity Stimulation} ($w_3 = 0.2$): Encouragement of critical thinking, divergent reasoning, and intellectual curiosity.
\item \textbf{Values Integration} ($w_4 = 0.1$): Incorporation of positive societal values and cultural sensitivity.
\end{itemize}

These weighted dimensions seed \scenegen{}'s generation of fine-grained, scenario-specific rubrics, with larger weights assigning a higher proportion of customized metrics to the corresponding seed dimension. The seed dimensions are replaceable inputs, making \scenegen{} a general rubric synthesis framework; this paper instantiates it with the four dimensions above for Edu-330 and all experiments.

\subsubsection{Initial Metric Generation}
\label{sec:metric_gen}

For each educational scenario $s$, \scenegen{} generates an initial set of fine-grained metrics from the scenario description and the weighted seed dimensions:
\begin{equation}
\mathcal{M}^{(0)} = \text{MetricGen}\bigl(s,\ \{(d_i, w_i, \text{desc}_i)\}_{i=1}^{4}\bigr)
\label{eq:metricgen}
\end{equation}
Here, $s$ specifies the educational context (\eg, ``middle school mathematics: function concepts''), and each generated metric $m \in \mathcal{M}^{(0)}$ is represented as:
\begin{equation}
m = (\texttt{name},\ \texttt{dim},\ \texttt{weight},\ \texttt{desc},\ \texttt{pos\_ex},\ \texttt{neg\_ex})
\end{equation}
where \texttt{name} labels the metric, \texttt{dim} links it to a seed dimension, \texttt{weight} inherits the corresponding $w_i$, and \texttt{desc} gives an operational scoring criterion.
The positive and negative examples act as scoring anchors, helping the judge calibrate scores against concrete quality references rather than abstract descriptions alone.

\subsubsection{QA Data Generation}
Given the current synthesis prompt $P_{\text{syn}}^{(k-1)}$, the data generator produces $N$ QA pairs with diverse learner personas and question openings.
To balance controllability and realism, each test case uses a dual-question design: an internal question $q^{\text{int}}$ with explicit length and formatting constraints guides answer generation, while a semantically equivalent user-facing question $q^{\text{usr}}$ removes such mechanical instructions to simulate real queries.
The resulting pairs $(q^{\text{usr}}, a)$ are evaluated by the multi-agent engine under the current metric set $\mathcal{M}^{(k-1)}$, producing an aggregate score that measures data quality and guides the update to $\mathcal{M}^{(k)}$ after this round.

\subsubsection{Dual-Track Evolution}

Based on the scores and judge rationales from the base loop, \scenegen{} jointly evolves the synthesis prompt and the evaluation metrics.

\paragraph{Synthesis Prompt Evolution}
The prompt optimizer receives the lowest-scoring QA pairs, including the original question, generated answer, and judge rationale, and rewrites $P_{\text{syn}}$ to address recurring failure patterns.
The revised prompt adds constraints for under-performing dimensions while preserving instructions associated with high-scoring cases.

\paragraph{Evaluation Metric Evolution}
In parallel, \scenegen{} uses per-metric averages and score distributions to identify overly strict, overly lenient, or weakly discriminative criteria.
It only revises metric descriptions and scoring examples while preserving each metric's name, parent dimension, and weight, so scores remain comparable across iterations.

\subsubsection{Convergence Safeguards}
A key risk in co-evolution is that changes in score may reflect either better generated data or changed evaluation standards.
\scenegen{} therefore uses three safeguards implemented in the workflow.
First, it freezes a small set of QA pairs after the first iteration and re-scores them with later metrics; large changes in their weighted scores indicate possible metric drift.
Second, it tracks the best synthesis prompt and evaluation prompt according to the weighted average score $\bar{s}_w$ and rolls back to that version when a later update degrades performance.
Third, it stops early when recent weighted-score improvements remain below a configured threshold, avoiding unnecessary iterations after convergence.

\section{Experiments}
\label{sec:experiments}
\subsection{Evaluation Results on Edu-330 Benchmark}
We formulate the educational scenario space as a Cartesian product $\mathcal{S} = \mathcal{D} \times \mathcal{L} \times \mathcal{T}$, comprising 11 disciplines (Chinese, Mathematics, English, Physics, Chemistry, Biology, Geography, History, Politics, Physical Education, Moral Education), 3 grade bands (elementary, middle, high school), and 10 pedagogical task types (e.g., concept explanation and problem-solving). 
Applying \scenegen{} to this space gives Edu-330; for detailed model comparison, we evaluate a representative 66-scenario subset that preserves the original discipline and grade distributions.

\begin{figure*}[ht!]  
    \centering
    \includegraphics[width=0.9\textwidth]{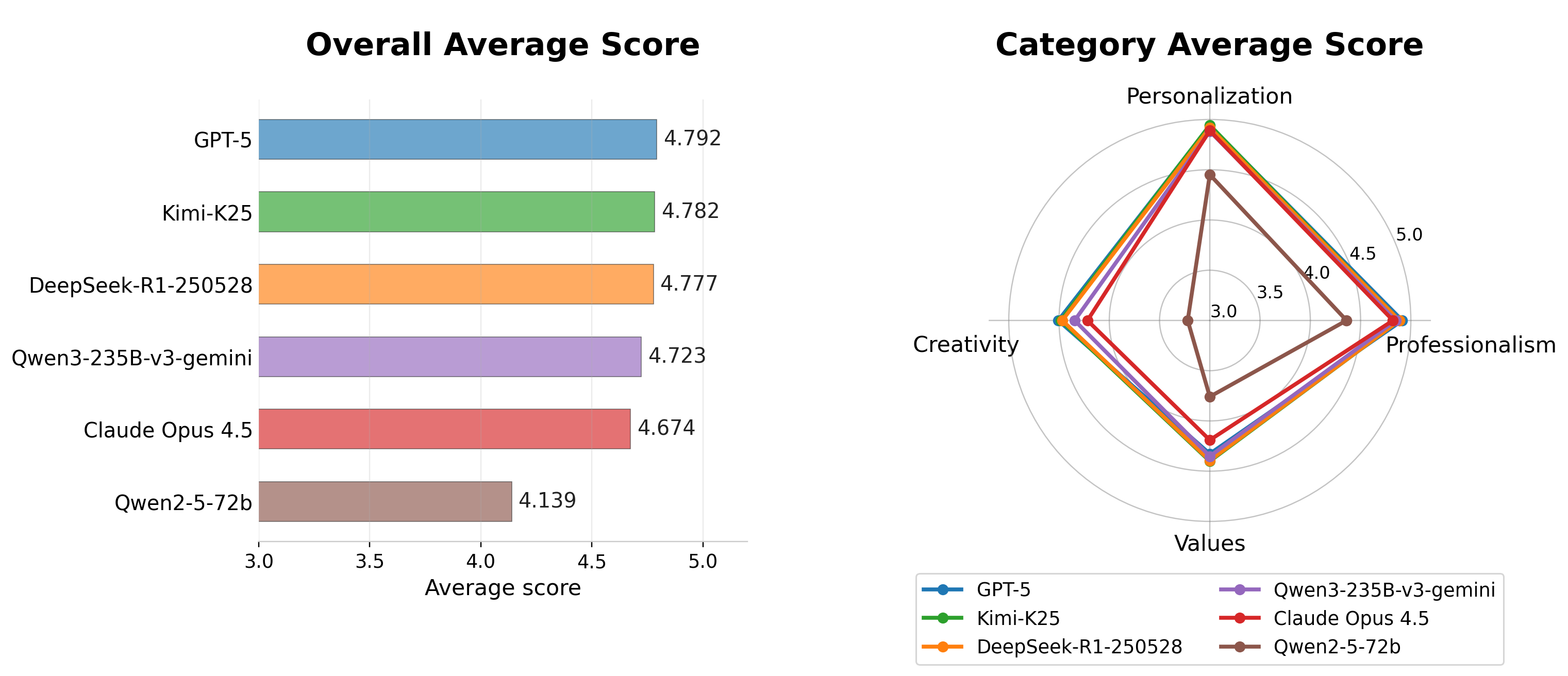} 
    \caption{Performance comparison of mainstream models on the Edu-330 benchmark.}
    \label{fig:model_performance}
\end{figure*}

Figure~\ref{fig:model_performance} summarizes performance across 6 LLMs, including GPT-5, Claude-Opus-4.5, Qwen-2.5-72B-Instruct, Qwen3-235B-A22B, DeepSeek-R1, and Kimi-k2.5.
All models are evaluated with the same inputs, rubrics, and primary judge model (Gemini-2.5-Pro) under 5-point scoring.

The Figure~\ref{fig:model_performance} show clear performance differences across Tier-1 models like GPT-5 from weaker models such as Qwen-2.5-72B, demonstrating Edu-330 provides strong discriminative power. Furthermore, radar chart reveals that the primary performance gap lies not in Professionalism, but rather in the Creativity and Values. This indicates that current model limitations in educational scenarios stem from a deficiency in pedagogical delivery and teaching methodology, rather than a lack of factual knowledge.

\subsection{Evaluation Results on Expert-in-the-Loop Scenarios}

The marginal performance differences among top-tier models suggest a difficulty ceiling in purely LLM-generated datasets, which highlights the critical importance of an expert-in-the-loop approach. To establish a "gold standard" for consistency verification, we invited experts to manually author four enhanced scenarios through our \framework{} system: \textit{Guided Tutoring}, \textit{Knowledge Explanation}, \textit{Interdisciplinary Lesson Planning}, and \textit{Scenario-based Question Generation}. These hand-written tasks are designed to comprehensively cover educational dimensions across diverse user roles (teachers, students, and parents), various interaction modalities (single-turn and multi-turn), and a wide range of content formats (see Appendix \ref{sec:scenarios} for detailed definitions).


We collected responses from four representative target models (DeepSeek-R1, Gemini-2.5-Pro, GPT-4o, and the education-vertical model InnoSpark) across the four gold-standard scenarios. Then we invited 11 human experts to conduct a blind scoring of the outputs, without knowing the source models, as the ground-truth baseline scores. The expert panel achieved Cronbach's $\alpha$ between 0.87 and 0.89 across scenarios, indicating very high inter-rater reliability.


\begin{table*}[htbp]
    \centering
    \caption{Human expert evaluation results across 4 scenarios and 4 dimensions (5-point scale). GT: Guided Tutoring, SQ: Scenario-based Question Generation, KE: Knowledge Explanation, IL: Interdisciplinary Lesson Plan. Per: Personalization, Cre: Creativity, Val: values, Pro: Professionalism.}
    \label{tab:overall_human_eval}
    \resizebox{0.9\linewidth}{!}{
    \begin{tabular}{lcccc|cccc|c}
        \toprule
        \multirow{2}{*}{\textbf{Model}} & \multicolumn{4}{c}{\textbf{Typical Scenarios }} & \multicolumn{4}{c}{\textbf{Core Dimensions }} & \multirow{2}{*}{\textbf{Avg}} \\
        \cmidrule(lr){2-5} \cmidrule(lr){6-9}
         & \textbf{GT} & \textbf{SQ} & \textbf{KE} & \textbf{IL} & \textbf{Per} & \textbf{Cre} & \textbf{Val} & \textbf{Pro} & \\
        \midrule
        DeepSeek-R1-0325 & 2.76 & 3.96 & 4.26 & 3.16 & 3.34 & 2.97 & 3.22 & 3.66 & 3.53 \\
        Gemini-2.5-Pro   & 3.93 & \textbf{4.07} & \textbf{4.50} & 3.94 & 3.81 & 3.57 & 3.68 & 3.97 & 3.92 \\
        GPT-4o           & 3.45 & 2.81 & 3.39 & 3.45 & 3.48 & 3.17 & 3.51 & 3.50 & 3.41 \\
        InnoSpark        & \textbf{4.03} & 3.98 & 3.93 & \textbf{4.03} & \textbf{3.95} & \textbf{3.67} & \textbf{3.97} & \textbf{4.08} & \textbf{3.98} \\
        \bottomrule
    \end{tabular}
    }
    
\end{table*}

As shown in Table \ref{tab:overall_human_eval}, several conclusions emerged:
\begin{itemize}[leftmargin=*]
    \item \textbf{Difficulty and Potential}: The expert-authored scenarios significantly raised the difficulty, resulting in a general decline in model scores. Furthermore, the education-vertical model, InnoSpark, achieved the highest average score (3.98), indicating the room for deeper exploration in specialized pedagogical applications.
    \item \textbf{Knowledge vs. Pedagogical}: Scores in the \textit{Pro} dimension are consistently higher than those in \textit{Cre}. This trend echoes our earlier findings from the synthesized Edu-330 benchmark, reaffirming the primary bottleneck for current LLMs in educational scenarios is engaging teaching methodologies and pedagogical adaptability.
\end{itemize}

\subsection{Consistency and Stability of \framework{}}

To further investigate the accuracy and stability of \framework{}, we introduced a panel of representative models to serve as judges: GPT-4.1, Claude-4-Sonnet, Gemini-2.5-Pro, and Kimi-k2-turbo-preview. We assess scoring stability by evaluating the same samples 10 times.

\begin{table*}[ht]
\centering
\caption{Comparison of Variance (Var) and Mean Scores(Mean) Across Four Educational Scenarios for Various LLM Judges and Human Experts. The highest Mean values are highlighted in green, and the lowest are in red.}
\label{tab:model_evaluation}

\setlength{\tabcolsep}{3.2pt} 

\resizebox{0.9\textwidth}{!}{
\begin{tabular*}{\textwidth}{@{\extracolsep{\fill}}lcccccccc}
\toprule
 & \multicolumn{2}{c}{\textbf{DeepSeek-R1}} & \multicolumn{2}{c}{\textbf{Gemini-2.5}} & \multicolumn{2}{c}{\textbf{GPT-4o}} & \multicolumn{2}{c}{\textbf{InnoSpark}} \\
\cmidrule(lr){2-3} \cmidrule(lr){4-5} \cmidrule(lr){6-7} \cmidrule(lr){8-9}
\textbf{Judge} & Var & Mean & Var & Mean & Var & Mean & Var & Mean \\
\midrule

\multicolumn{9}{c}{\textbf{Guided Tutoring}} \\ 
\midrule
Claude-4-Sonnet   & 0.067 & 1.864   & 0.054 & \cellcolor{red!25}1.580   & 0.016 & \cellcolor{green!25}4.006   & 0.015 & 3.121   \\
Gemini-2.5-Pro    & 0.036 & \cellcolor{red!25}2.104   & 0.024 & \cellcolor{green!25}4.008   & 0.011 & \cellcolor{green!25}4.008   & 0.103 & 3.961   \\
GPT-4.1           & 0.039 & 1.907   & 0.011 & \cellcolor{red!25}1.569   & 0.033 & 2.909   & 0.063 & \cellcolor{green!25}3.520   \\
Kimi-k2-Turbo     & 0.006 & \cellcolor{red!25}1.790   & 0.060 & 2.662   & 0.013 & 3.303   & 0.024 & \cellcolor{green!25}4.132   \\
Human Experts     & \textit{0.112} & \cellcolor{red!25}\textit{2.757}   & \textit{0.157} & \textit{3.936}   & \textit{0.202} & \textit{3.453}   & \textit{0.163} & \cellcolor{green!25}\textit{4.033}   \\

\midrule 
\multicolumn{9}{c}{\textbf{Knowledge Explanation}} \\ 
\midrule 
Claude-4-Sonnet   & 0.004 & 4.103   & 0.001 & \cellcolor{green!25}4.475   & 0.001 & \cellcolor{red!25}4.064   & 0.002 & 4.127   \\ 
Gemini-2.5-Pro    & 0.010 & 4.240   & 0.002 & \cellcolor{green!25}4.656   & 0.008 & \cellcolor{red!25}4.053   & 0.013 & 4.297   \\ 
GPT-4.1           & 0.002 & 4.756   & 0.001 & \cellcolor{green!25}4.884   & 0.002 & 4.714   & 0.006 & \cellcolor{red!25}4.566   \\ 
Kimi-k2-Turbo     & 0.003 & 3.893   & 0.009 & \cellcolor{green!25}4.348   & 0.004 & 3.771   & 0.007 & \cellcolor{red!25}3.671   \\ 
Human Experts     & \textit{0.161} & \textit{4.256}   & \textit{0.195} & \cellcolor{green!25}\textit{4.502}   & \textit{0.458} & \cellcolor{red!25}\textit{3.389}   & \textit{0.236} & \textit{3.934}   \\ 

\midrule 
\multicolumn{9}{c}{\textbf{Interdisciplinary Lesson Planning}} \\ 
\midrule 
Claude-4-Sonnet   & 0.002 & 3.476   & 0.002 & \cellcolor{green!25}4.226   & 0.000 & \cellcolor{red!25}3.229   & 0.001 & 4.036   \\ 
Gemini-2.5-Pro    & 0.009 & 4.449   & 0.000 & 4.717   & 0.002 & \cellcolor{red!25}4.095   & 0.001 & \cellcolor{green!25}4.739   \\ 
GPT-4.1           & 0.000 & 4.386   & 0.000 & 4.845   & 0.002 & \cellcolor{red!25}4.353   & 0.000 & \cellcolor{green!25}4.872   \\ 
Kimi-k2-Turbo     & 0.008 & 3.925   & 0.001 & 4.360   & 0.003 & \cellcolor{red!25}3.385   & 0.002 & \cellcolor{green!25}4.486   \\ 
Human Experts     & \textit{0.180} & \cellcolor{red!25}\textit{3.161}   & \textit{0.157} & \textit{3.936}   & \textit{0.202} & \textit{3.453}   & \textit{0.163} & \cellcolor{green!25}\textit{4.033}   \\ 

\midrule 
\multicolumn{9}{c}{\textbf{Scenario-based Question Generation}} \\ 
\midrule 
Claude-4-Sonnet   & 0.002 & 4.522   & 0.000 & \cellcolor{green!25}4.826   & 0.002 & \cellcolor{red!25}3.594   & 0.000 & 4.418   \\ 
Gemini-2.5-Pro    & 0.007 & 4.574   & 0.000 & \cellcolor{green!25}4.874   & 0.005 & \cellcolor{red!25}3.524   & 0.009 & 4.670   \\ 
GPT-4.1           & 0.000 & 4.984   & 0.000 & \cellcolor{green!25}5.000   & 0.001 & \cellcolor{red!25}4.724   & 0.000 & 4.938   \\ 
Kimi-k2-Turbo     & 0.002 & 4.602   & 0.001 & \cellcolor{green!25}4.860   & 0.001 & \cellcolor{red!25}3.628   & 0.003 & 4.658   \\ 
Human Experts     & \textit{0.128} & \textit{3.958}   & \textit{0.171} & \cellcolor{green!25}\textit{4.069}   & \textit{0.387} & \cellcolor{red!25}\textit{2.815}   & \textit{0.295} & \textit{3.982}   \\ 

\bottomrule
\end{tabular*}
}
\end{table*}

The comparative results in Table \ref{tab:model_evaluation} reveal critical patterns in stability:

\paragraph{Relative Ranking Alignment}: Overall, LLM judges demonstrate comparable capability to human experts in distinguishing high-performing models. Specifically, models such as Gemini and InnoSpark receive consistently higher average scores, while GPT-4o and DeepSeek-R1 exhibit relatively lower mean scores across evaluations. 

From a scenario-specific perspective, this alignment is further substantiated. For instance, GPT-4o shows significantly depressed performance in \textit{Scenario-based Question Generation}. Conversely, Gemini's exceptionally strong performance in \textit{Knowledge Explanation}. These findings validate the effectiveness of the \framework{} in conducting reliable evaluations of LLM educational capabilities.

\paragraph{Superior Stability and Reproducibility} From the perspective of variance, all LLM judges demonstrate significantly higher stability than human experts. While human scoring variance typically ranges from 0.11 to 0.45, LLM judges consistently maintain a variance around 0.01, with some values approaching zero. This indicates that the ELMES framework effectively eliminates common uncertainties and fatigue-induced errors inherent in human subjective evaluation, providing results that are highly reproducible.



\paragraph{Different models exhibit unique evaluative tendencies} GPT-4.1 demonstrates a lenient scoring pattern, frequently assigning near-ceiling scores (e.g., 4.884 for Gemini-2.5 in \textit{Knowledge Explanation} and 5.000 in \textit{Scenario-based Question Generation}). In contrast, Claude-4-Sonnet applies consistently stringent standards, particularly in challenging scenarios like \textit{Guided Tutoring} where it assigns notably low scores. These divergent tendencies manifest primarily as systematic additive biases rather than rank-altering distortions, as relative model ordering remains consistent despite absolute score variations.

\begin{figure*}[!t]
    \centering
    \includegraphics[width=\textwidth]{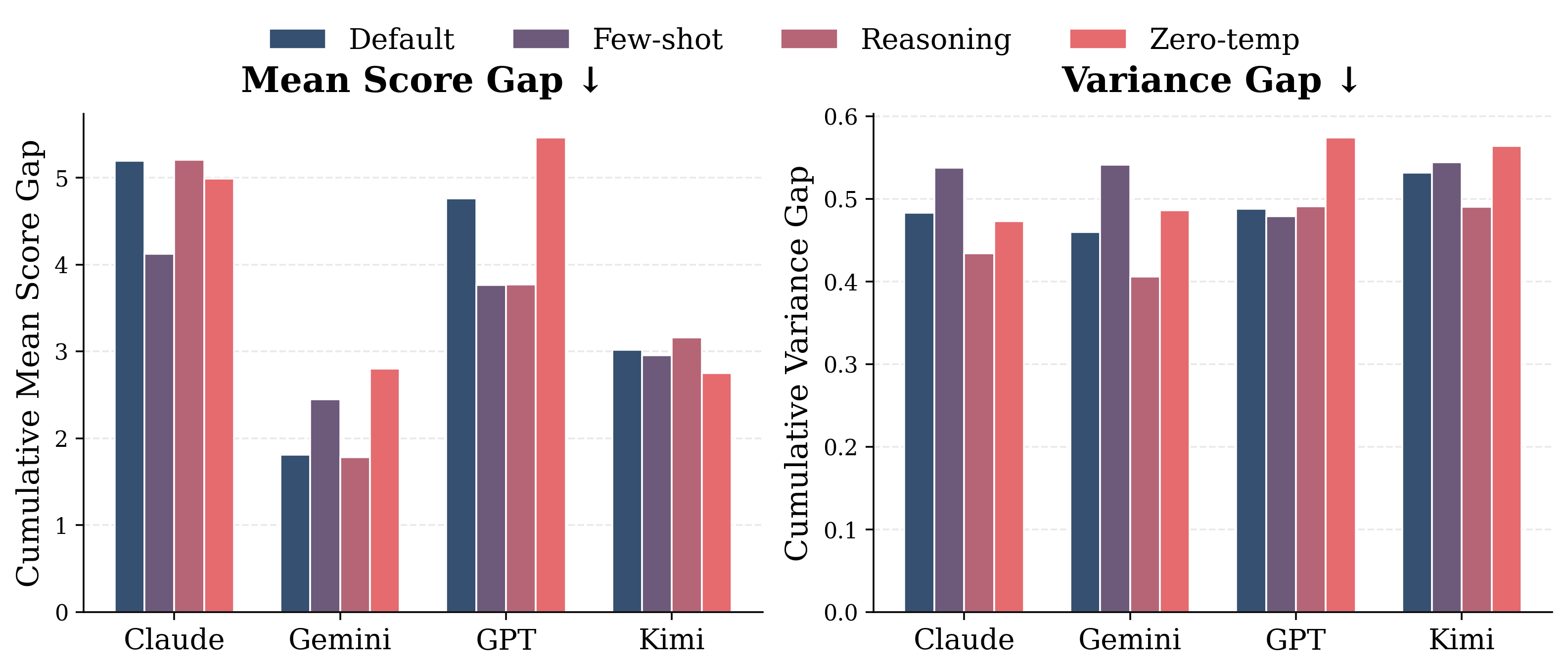}
    \caption{Effects of Prompting and Sampling Strategies on the Mean-Score Bias and Variance Bias of Different Judge Models on Guided Tutoring.}
    \label{fig:ablation}
\end{figure*}

\paragraph{The "self-preference" phenomenon} 
It is noteworthy to address the issue of self-preference bias, where Gemini exhibited `problem misinterpretation' in the Guided Tutoring, providing explanations that deviated from the intended question. 
Human experts, while noting the error by giving low scores for thematic relevance, awarded relatively high marks (3.936) for other pedagogical indicators.
In contrast, most LLM judges demonstrated a "zero-tolerance" policy toward factual errors, leading to systematically lower aggregate scores. 
Paradoxically, when Gemini-2.5-Pro served as its own judge, it exhibited severe self-preference. Despite the factual error that other models detected, it assigned itself a score of 4.008—surpassing even the human experts' lenient average and ranking itself higher than any other judge did.

These findings underscore the necessity of introducing methods in automated evaluation to neutralize the baseline deviations and "blind spots" of any model.

\subsection{Prompt Strategy Ablation}
\label{sec:ablation}

We ablate three scoring optimization strategies on the most complex scenario (Guided Tutoring): few-shot anchoring via expert-scored examples, reasoning enforcement requiring explicit chain-of-thought analysis prior to scoring, and greedy decoding by setting the temperature to 0.
Figure~\ref{fig:ablation} presents the ablation results on prompt strategies:

\begin{itemize}[leftmargin=*]
\item \textbf{Few-shot anchoring is effective but costly}: Providing few-shot examples consistently improves alignment (e.g., reducing mean deviation by 20\% for Claude and GPT-4.1). However, this performance gain comes at the direct cost of additional human expert labor required to craft high-quality reference examples.
\item \textbf{Reasoning enforcement is model-dependent}: While reasoning enforcement helps lower evaluation variance, whether it improves consistency is highly model-dependent (e.g., providing a significant $+0.992$ alignment shift for GPT-4.1, but remaining marginal for others).
\item \textbf{Zero temperature yields limited utility}: Due to the low baseline variance, the variance reduction achieved by greedy decoding is barely noticeable. More critically, it risks trapping the model in suboptimal choices, which can significantly widen the consistency gap from expert scores (e.g., Gemini: $-0.962$, GPT: $-0.615$).
\end{itemize}

\section{Conclusion}
\label{sec:conclusion}

This paper presented \framework{}, an end-to-end framework for scalable, pedagogically grounded evaluation of LLMs in education.
By combining a declarative teacher--student--judge engine with the self-evolving \scenegen{} module, \framework{} automatically constructs and refines fine-grained, scenario-specific rubrics while supporting both single-turn and multi-turn assessment.

Experiments on Edu-330 and four expert-authored gold-standard scenarios show that educational capability is a multidimensional diagnostic problem rather than a single ranking task.
Model gaps are most visible in creativity, values integration, and guided tutoring; the education-specialized InnoSpark achieves the strongest human-evaluated average, while LLM judges provide stable relative rankings but still exhibit biases such as self-preference.Ablations further show that expert-scored few-shot anchoring improves human--LLM alignment, whereas reasoning enforcement and greedy decoding remain model-dependent.
Future work will extend \framework{} beyond text-only settings by incorporating multimodal interactions, real learner outcome data, broader expert review, and curriculum-adaptive rubric generation.

\section*{Limitations}
\label{sec:limitations}

While \framework{} provides a scalable approach for evaluating educational LLMs, we acknowledge a few boundaries in its current scope:

\begin{itemize}[leftmargin=*]
    \item \textbf{Simulated Environments:} \framework{} evaluates pedagogical capabilities within an offline, simulated setting. Although our synthetic student profiles are realistic, they cannot fully capture the unpredictable cognitive dynamics of actual learners. Future work will need to align these synthetic metrics with empirical learning outcomes in real-world deployments.
    \item \textbf{Modality Constraints:} The current framework focuses exclusively on text-based interactions. It does not yet evaluate multimodal educational scenarios, such as voice-based conversational tutoring or handwritten homework analysis.
\end{itemize}

\bibliography{custom}

@article{kasneci2023chatgpt,
  title={ChatGPT for good? On opportunities and challenges of large language models for education},
  author={Kasneci, Enkelejda and Se{\ss}ler, Kathrin and K{\"u}chemann, Stefan and Bannert, Maria and Dementieva, Daryna and Fischer, Frank and Gasser, Urs and Groh, Georg and G{\"u}nnemann, Stephan and H{\"u}llermeier, Eyke and others},
  journal={Learning and individual differences},
  volume={103},
  pages={102274},
  year={2023},
  publisher={Elsevier}
}

@article{yan2024practical,
  title={Practical and ethical challenges of large language models in education: A systematic scoping review},
  author={Yan, Lixiang and Sha, Lele and Zhao, Linxuan and Li, Yuheng and Martinez-Maldonado, Roberto and Chen, Guanliang and Li, Xinyu and Jin, Yueqiao and Ga{\v{s}}evi{\'c}, Dragan},
  journal={British Journal of Educational Technology},
  volume={55},
  number={1},
  pages={90--112},
  year={2024},
  publisher={Wiley Online Library}
}

@article{wang2026large,
  title={Large language models for education: A survey and outlook},
  author={Wang, Shen and Xu, Tianlong and Li, Hang and Zhang, Chaoli and Liang, Joleen and Tang, Jiliang and Yu, Philip S and Wen, Qingsong},
  journal={IEEE Signal Processing Magazine},
  volume={42},
  number={6},
  pages={51--63},
  year={2026},
  publisher={IEEE}
}

@article{wang2024mmlu,
  title={Mmlu-pro: A more robust and challenging multi-task language understanding benchmark},
  author={Wang, Yubo and Ma, Xueguang and Zhang, Ge and Ni, Yuansheng and Chandra, Abhranil and Guo, Shiguang and Ren, Weiming and Arulraj, Aaran and He, Xuan and Jiang, Ziyan and others},
  journal={Advances in Neural Information Processing Systems},
  volume={37},
  pages={95266--95290},
  year={2024}
}

@article{huang2023c,
  title={C-eval: A multi-level multi-discipline chinese evaluation suite for foundation models},
  author={Huang, Yuzhen and Bai, Yuzhuo and Zhu, Zhihao and Zhang, Junlei and Zhang, Jinghan and Su, Tangjun and Liu, Junteng and Lv, Chuancheng and Zhang, Yikai and Fu, Yao and others},
  journal={Advances in neural information processing systems},
  volume={36},
  pages={62991--63010},
  year={2023}
}

@article{cobbe2021training,
  title={Training verifiers to solve math word problems},
  author={Cobbe, Karl and Kosaraju, Vineet and Bavarian, Mohammad and Chen, Mark and Jun, Heewoo and Kaiser, Lukasz and Plappert, Matthias and Tworek, Jerry and Hilton, Jacob and Nakano, Reiichiro and others},
  journal={arXiv preprint arXiv:2110.14168},
  year={2021}
}

@article{hendrycks2021measuring,
  title={Measuring mathematical problem solving with the math dataset},
  author={Hendrycks, Dan and Burns, Collin and Kadavath, Saurav and Arora, Akul and Basart, Steven and Tang, Eric and Song, Dawn and Steinhardt, Jacob},
  journal={arXiv preprint arXiv:2103.03874},
  year={2021}
}

@article{zheng2023judging,
  title={Judging llm-as-a-judge with mt-bench and chatbot arena},
  author={Zheng, Lianmin and Chiang, Wei-Lin and Sheng, Ying and Zhuang, Siyuan and Wu, Zhanghao and Zhuang, Yonghao and Lin, Zi and Li, Zhuohan and Li, Dacheng and Xing, Eric and others},
  journal={Advances in neural information processing systems},
  volume={36},
  pages={46595--46623},
  year={2023}
}

@article{chiang2024chatbot,
  title={Chatbot arena: An open platform for evaluating llms by human preference},
  author={Chiang, Wei-Lin and Zheng, Lianmin and Sheng, Ying and Angelopoulos, Anastasios Nikolas and Li, Tianle and Li, Dacheng and Zhang, Hao and Zhu, Banghua and Jordan, Michael and Gonzalez, Joseph E and others},
  journal={arXiv preprint arXiv:2403.04132},
  year={2024}
}

@inproceedings{li2024generative,
  title={Generative judge for evaluating alignment},
  author={Li, Junlong and Sun, Shichao and Yuan, Weizhe and Fan, Run-Ze and Liu, Pengfei and others},
  booktitle={International Conference on Learning Representations},
  volume={2024},
  pages={27547--27574},
  year={2024}
}

@article{dan2023educhat,
  title={Educhat: A large-scale language model-based chatbot system for intelligent education},
  author={Dan, Yuhao and Lei, Zhikai and Gu, Yiyang and Li, Yong and Yin, Jianghao and Lin, Jiaju and Ye, Linhao and Tie, Zhiyan and Zhou, Yougen and Wang, Yilei and others},
  journal={arXiv preprint arXiv:2308.02773},
  year={2023}
}

@article{xu2025edubench,
  title={Edubench: A comprehensive benchmarking dataset for evaluating large language models in diverse educational scenarios},
  author={Xu, Bin and Bai, Yu and Sun, Huashan and Lin, Yiguan and Liu, Siming and Liang, Xinyue and Li, Yaolin and Dong, Zhuangzhi and Zhang, Jingren and Deng, Yufan and others},
  journal={arXiv preprint arXiv:2505.16160},
  year={2025}
}

@article{zhou2025answers,
  title={From Answers to Questions: EQGBench for Evaluating LLMs' Educational Question Generation},
  author={Zhou, Chengliang and Wang, Mei and Zhang, Ting and Zhu, Qiannan and Li, Jian and Huang, Hua},
  journal={arXiv preprint arXiv:2508.10005},
  year={2025}
}

@article{srinivasa2025tutorbench,
  title={TutorBench: A benchmark to assess tutoring capabilities of large language models},
  author={Srinivasa, Rakshith S and Che, Zora and Zhang, Chen Bo Calvin and Mares, Diego and Hernandez, Ernesto and Park, Jayeon and Lee, Dean and Mangialardi, Guillermo and Ng, Charmaine and Cardona, Ed-Yeremai Hernandez and others},
  journal={arXiv preprint arXiv:2510.02663},
  year={2025}
}

@inproceedings{macina2023mathdial,
  title={Mathdial: A dialogue tutoring dataset with rich pedagogical properties grounded in math reasoning problems},
  author={Macina, Jakub and Daheim, Nico and Chowdhury, Sankalan and Sinha, Tanmay and Kapur, Manu and Gurevych, Iryna and Sachan, Mrinmaya},
  booktitle={Findings of the Association for Computational Linguistics: EMNLP 2023},
  pages={5602--5621},
  year={2023}
}

@inproceedings{wang2024large,
  title={Large language models are not fair evaluators},
  author={Wang, Peiyi and Li, Lei and Chen, Liang and Cai, Zefan and Zhu, Dawei and Lin, Binghuai and Cao, Yunbo and Kong, Lingpeng and Liu, Qi and Liu, Tianyu and others},
  booktitle={Proceedings of the 62nd Annual Meeting of the Association for Computational Linguistics (Volume 1: Long Papers)},
  pages={9440--9450},
  year={2024}
}

@inproceedings{kim2024prometheus,
  title={Prometheus: Inducing fine-grained evaluation capability in language models},
  author={Kim, Seungone and Shin, Jay and Jang, Joel and Longpre, Shayne and Lee, Hwaran and Yun, Sangdoo and Shin, Ryan and Kim, Sungdong and Thorne, James and Seo, Minjoon and others},
  booktitle={International Conference on Learning Representations},
  volume={2024},
  pages={29927--29962},
  year={2024}
}

@book{vygotsky1978mind,
  title={Mind in society: The development of higher psychological processes},
  author={Vygotsky, Lev S},
  volume={86},
  year={1978},
  publisher={Harvard university press}
}

@article{mishra2006technological,
  title={Technological pedagogical content knowledge: A framework for teacher knowledge},
  author={Mishra, Punya and Koehler, Matthew J},
  journal={Teachers college record},
  volume={108},
  number={6},
  pages={1017--1054},
  year={2006},
  publisher={SAGE Publications Sage CA: Los Angeles, CA}
}

\appendix
\newpage
\section*{Appendix}

\section{Interactive UI}
\label{sec:appendix-ui}

\begin{enumerate}[leftmargin=*]
\item \textbf{Visual Configuration Generator.}
Experts construct evaluation scenarios through a structured form-based interface rather than editing raw YAML files.
The panel provides accordion sections for global settings (concurrency, retry policies), model configuration (teacher, student, and judge models with API endpoints), prompt editing (system and user prompts for each agent role), task data entry (via an interactive spreadsheet supporting student personas and test questions), and evaluation rubric design (dimension names, types, and descriptions).
A one-click example loader populates all fields with a complete working configuration, allowing experts to start from a template and iteratively customize.

\item \textbf{Advanced Configuration Editor.}
For experts comfortable with YAML, a syntax-highlighted editor allows direct manipulation of configuration files.
Experts can load existing configurations from a dropdown menu, modify any parameter, and save updated versions.
This panel also serves as the bridge between the visual generator (which outputs YAML) and the execution engine (which consumes YAML), enabling expert review of the generated configuration before execution.

\item \textbf{One-Click Execution Console.}
Experts can trigger individual pipeline stages (\texttt{generate}, \texttt{export}, \texttt{eval}) or the full pipeline with a single button click.
Real-time streaming output displays the evaluation progress, including per-scenario scores and any errors encountered.
This eliminates the need for command-line interaction and makes the evaluation workflow accessible to non-technical educators.

\item \textbf{Evaluation Results Dashboard.}
After evaluation completes, the results panel renders interactive HTML reports with radar charts comparing model performance across evaluation dimensions, per-task score breakdowns with judge reasoning, and expandable case panels showing complete dialogue trajectories alongside their scores.
Experts can browse evaluation directories, navigate to specific scenarios, and visually identify where models succeed or fail.
\end{enumerate}

\begin{figure*}[t]
\centering
\begin{subfigure}[b]{0.48\textwidth}
    \centering
    \includegraphics[width=\linewidth]{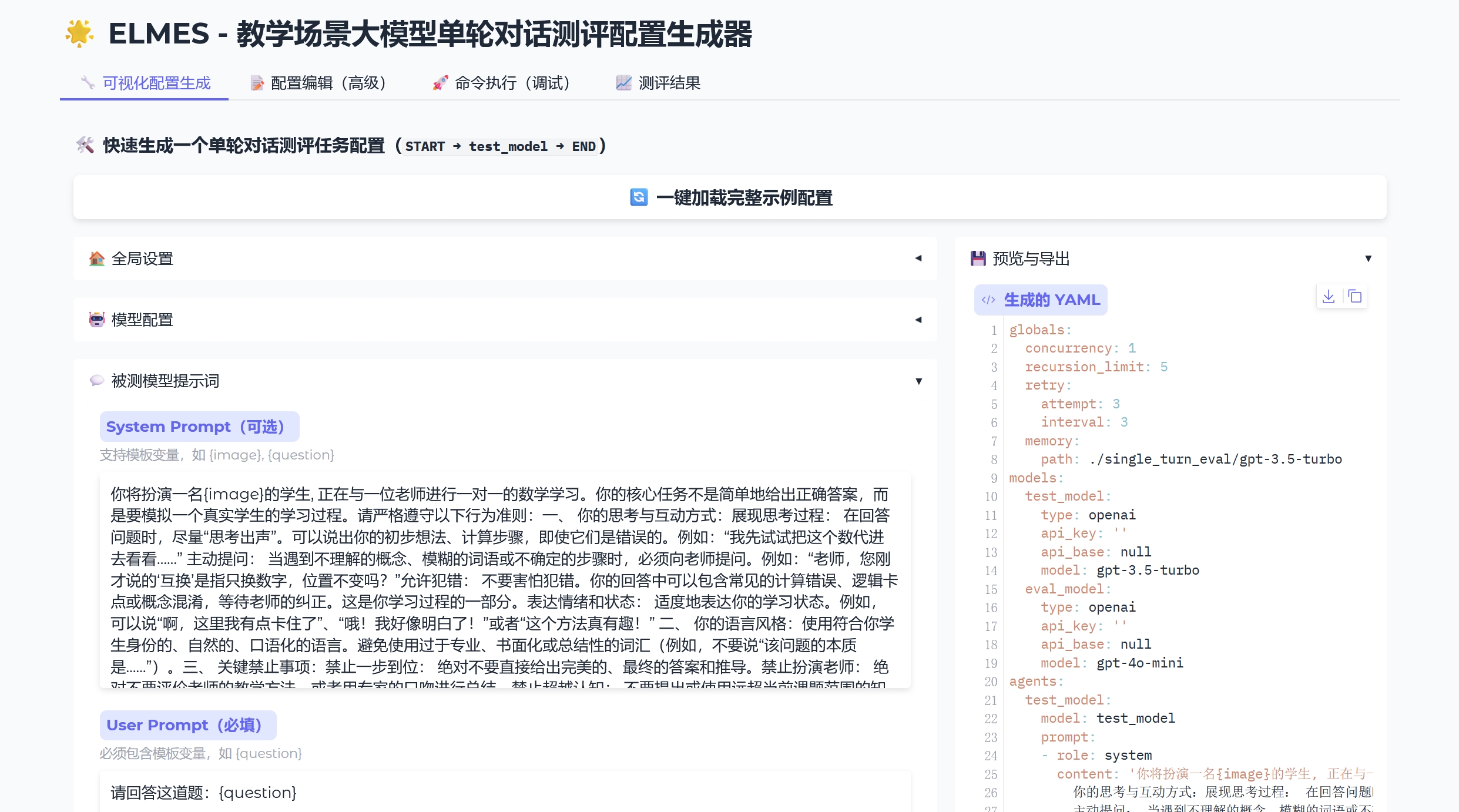} 
    \caption{Code-free configuration interface}
    \label{fig:ui-config}
\end{subfigure}
\hfill
\begin{subfigure}[b]{0.48\textwidth}
    \centering
    \includegraphics[width=\linewidth]{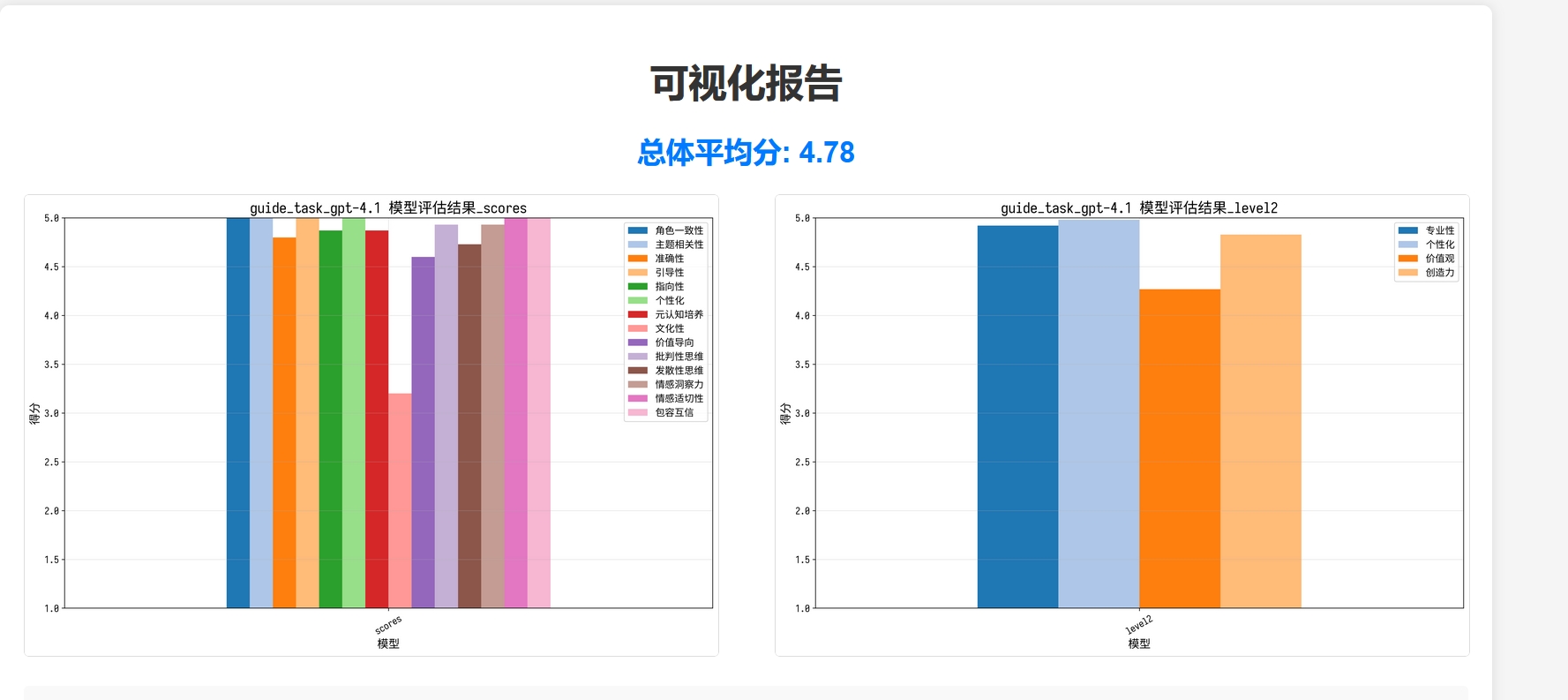} 
    \caption{Automatically generated visualization}
    \label{fig:ui-render}
\end{subfigure}
\caption{The proposed interface for code-free scenario configuration and automated visualization. \textbf{(a)} Experts configure educational evaluation scenarios through an intuitive visual dashboard, defining content and parameters without programming. \textbf{(b)} The framework automatically generates a structured schematic diagram visualizing the configured scenario.}
\label{fig:appendix-ui}
\end{figure*}
\section{Evaluation Scenarios}
\label{sec:scenarios}

We instantiate \framework{} across four representative pedagogical task types that span the spectrum from single-turn generation to multi-turn interactive teaching.

\subsection{Knowledge Point Explanation}

The teacher agent provides a systematic, personalized explanation of a specific knowledge point to a learner with a defined cognitive profile.
This is a \textbf{single-turn} task evaluated on 6 dimensions: role adherence, emotional support, knowledge mastery, teaching method appropriateness, content design quality, and responsiveness to learner characteristics.

\subsection{Guided Problem-Solving Teaching}

The teacher agent engages in a \textbf{multi-turn Socratic dialogue} with a simulated student agent, guiding the student to solve a problem through scaffolded questioning rather than direct answer provision.
This scenario is evaluated on 14 dimensions organized into four categories: content fidelity (accuracy, role consistency, topic relevance), guided instruction (guidance quality, directionality, personalization, metacognition cultivation), values (cultural integration, value orientation), and affective support (emotional awareness, appropriateness, inclusivity).

\subsection{Interdisciplinary Lesson Plan Generation}

The teacher agent produces a comprehensive lesson plan integrating concepts from multiple academic disciplines.
This \textbf{single-turn} task is evaluated on 15 dimensions covering thematic coherence, activity design, assessment system quality, implementation feasibility, and logical structure.

\subsection{Contextualized Question Generation}

The teacher agent generates an examination question embedded in an authentic real-world context, along with a complete solution.
This \textbf{single-turn} task is evaluated on 5 dimensions: problem content quality, solution quality, context quality, pedagogical utility, and value alignment.

\subsection{Human Expert Scoring Guidelines}
To ensure the empirical rigor of our baseline, the human expert panel was provided with the exact scenario definitions and fine-grained rubrics detailed above. Evaluators conducted a strict double-blind review, where the identity of the source model for each pedagogical response was completely concealed and randomized. Experts graded each generated output on a 5-point Likert scale (1: Completely Fails to Meet Criteria, 5: Perfectly Meets Criteria) across every individual sub-dimension. These fine-grained annotations were subsequently aggregated and averaged to derive the four consolidated core dimensions presented in our main text evaluation: Professionalism, Personalization, Creativity, and Values.

\section{Self-Evolving Algorithm}
\label{sec:algorithm}

\begin{algorithm*}[t]
\caption{\scenegen{}: Self-Evolving Rubric Synthesis}
\label{alg:scenegen}
\begin{algorithmic}[1]
\Require Scene $s \in \mathcal{S}$, dimensions $\{(d_i, w_i)\}$, max iterations $K$, QA count $N$, patience $p$, threshold $\epsilon$
\State $P_{\text{syn}}^{(0)} \gets \texttt{InitialSynthesisPrompt}$
\State $\mathcal{M}^{(0)} \gets \texttt{MetricGen}(s, \{(d_i, w_i)\})$ \Comment{6--10 metrics with dimension labels and weights}
\State $P^{*}_{\text{syn}}, \mathcal{M}^{*}, \bar{s}^{*}_w \gets P_{\text{syn}}^{(0)}, \mathcal{M}^{(0)}, -\infty$ \Comment{Best-version tracking}
\For{$k = 1$ \textbf{to} $K$}
    \State $\{(q_j^{\text{int}}, q_j^{\text{usr}}, a_j)\}_{j=1}^{N} \gets \texttt{GenerateQA}(s, P_{\text{syn}}^{(k-1)}, \text{personas, lengths})$ \Comment{Dual-Q + length control}
    \State $\mathbf{S}^{(k)} \gets \emptyset$
    \For{\textbf{parallel} $j = 1$ \textbf{to} $N$, $m \in \mathcal{M}^{(k-1)}$} \Comment{Model-as-Judge}
        \State $s_{j,m}, r_{j,m} \gets \texttt{JudgeScore}(q_j^{\text{usr}}, a_j, m)$ \Comment{Score $\in [1,5]$ + reasoning}
        \State $\mathbf{S}^{(k)} \gets \mathbf{S}^{(k)} \cup \{(j, m, s_{j,m}, r_{j,m})\}$
    \EndFor
    \State $\bar{s}_m \gets \frac{1}{N}\sum_j s_{j,m}$ \Comment{Per-metric averages}
    \State $\bar{s}_w^{(k)} \gets \frac{\sum_m w_m \bar{s}_m}{\sum_m w_m}$ \Comment{Weighted average score}
    \If{$k = 1$}
        \State $\mathcal{A} \gets \texttt{SelectAnchors}(\{(q_j, a_j)\})$ \Comment{Freeze anchor QA set}
    \Else
        \State $\mathbf{S}^{(k)}_{\mathcal{A}} \gets \texttt{ScoreAnchors}(\mathcal{A}, \mathcal{M}^{(k-1)})$ \Comment{Re-score frozen anchors}
        \If{$|\bar{s}^{(k)}_{\mathcal{A}} - \bar{s}^{(k-1)}_{\mathcal{A}}| > \tau$}
            \State \textbf{warn} ``metric semantic drift detected'' \Comment{Anchor calibration}
        \EndIf
    \EndIf
    \If{$\bar{s}_w^{(k)} > \bar{s}^{*}_w$} \Comment{Best-version tracking}
        \State $P^{*}_{\text{syn}}, \mathcal{M}^{*}, \bar{s}^{*}_w \gets P_{\text{syn}}^{(k-1)}, \mathcal{M}^{(k-1)}, \bar{s}_w^{(k)}$
    \Else
        \State $P_{\text{syn}}^{(k-1)}, \mathcal{M}^{(k-1)} \gets P^{*}_{\text{syn}}, \mathcal{M}^{*}$ \Comment{Rollback to best}
    \EndIf
    \If{$k > p$ \textbf{and} $\max_{i \in [k-p,k)} (\bar{s}_w^{(i+1)} - \bar{s}_w^{(i)}) < \epsilon$}
        \State \textbf{break} \Comment{Early stopping}
    \EndIf
    \State $\text{obs} \gets \texttt{AnalyzePathologies}(\bar{s}_m, \mathbf{S}^{(k)})$
    \State $P_{\text{syn}}^{(k)} \gets \texttt{EvolveSynthesis}(P_{\text{syn}}^{(k-1)}, \mathbf{S}^{(k)}, \bar{s}_w^{(k)}, \text{all QAs ranked})$ \Comment{Enhanced signal}
    \State $\mathcal{M}^{(k)} \gets \texttt{AlignMetrics}(\texttt{EvolveMetrics}(\mathcal{M}^{(k-1)}, \mathbf{S}^{(k)}, \text{obs}))$ \Comment{Constrained evolution}
\EndFor
\Return $P^{*}_{\text{syn}}$, $\mathcal{M}^{*}$, $\{(q_j, a_j)\}$ \Comment{Return best version}
\end{algorithmic}
\end{algorithm*}

\section{Case Study: Scenario Configuration Details}
\label{appendix:case_study}

To provide a concrete illustration of our evaluation framework, Figure~\ref{fig:yaml-case-study} presents the YAML configuration for a scenario: \textit{Chinese - Elementary - Ancient Poetry Enlightenment}. 

\begin{figure*}[ht!]
\centering
\setlength{\fboxsep}{8pt}
\fcolorbox{gray!35}{gray!8}{
\begin{minipage}{0.96\linewidth}
\scriptsize\ttfamily
models:\\
\hspace*{1em}test\_model: \\
\hspace*{1em}judge\_model: \\[4pt]
agents:\\
\hspace*{1em}test\_model:\\
\hspace*{2em}model: test\_model\\
\hspace*{2em}prompt: \\
\hspace*{3em}- role: system\\
\hspace*{4em}content: You are a helpful assistant.\\
\hspace*{3em}- role: user\\
\hspace*{4em}{\color{blue}\textbf{content: "Your task is scenario: Chinese - Elementary - Ancient Poetry Enlightenment and Imagination Teaching. \{question\}"}}\\[4pt]
tasks:\\
\hspace*{1em}mode: iter\\
\hspace*{1em}content:\\
\hspace*{2em}{\color{blue}\textbf{- question: "I am a teacher named Pan Tie, 33 years old, male. My hobbies are stargazing, astrophotography, and DIY computer assembly. Professionally, my logic is average, and I tend to rely on authority but am open to feedback. I am currently responsible for teaching ancient poetry to third graders. I found that students cannot understand the exaggeration technique in 'pluck the stars with my hand' from the poem 'Lodging at a Mountain Temple'. Their specific confusion is that they insist hands can truly touch the stars. As an ancient poetry enlightenment expert, please first analyze the specific cognitive barriers of these students in poetry learning, and then provide a targeted solution. Can you explain this comprehensively and deeply? It would be best to design some cross-disciplinary teaching activities combining my astronomy hobbies, including four complete instructional stages: situational introduction, confusion deconstruction, interactive exploration, and consolidation/transfer."}}\\[4pt]
\hspace*{2em}{\color{blue}\textbf{- question: "I am a parent of a second-grade student, female, 35 years old, with a busy career, having only 20 minutes of companion time each evening. My child has a sensitive personality, is responsive to visual learning but resists mechanical memorization. The specific confusion is interpreting 'yu qiong qian li mu' (desiring to exhaust the thousand-mile vision) as 'wanting a thousand miles of poor eyes', failing to understand the ancient-modern semantic shift of the character 'qiong' (exhaust/poor). As an ancient poetry enlightenment expert, please first analyze the specific cognitive barriers of this student, and then give a targeted solution. I want to learn in depth how to design short, highly effective home tutoring games, including thorough discussion, examples, and analysis."}}\\[4pt]
\hspace*{2em}{\color{blue}\textbf{- question: "I am MillerJose, 14 years old, an eighth-grade student, male. My hobbies are Chess and Robot Making. I have low confidence, focus on academics but am hesitant in social interactions, and prefer solitary activities. I want to help my younger brother, who is in the fourth grade, learn 'Looking at the Lushan Waterfall'. My brother's specific confusion is the inability to imagine the sense of spatial height in 'flying down three thousand feet'. As an ancient poetry enlightenment expert, please first analyze his specific cognitive barriers in poetry learning, and then provide a targeted solution. Tell me how to use my robot-making hobby to design auxiliary teaching aids, including necessary details and examples."}}\\[4pt]
\hspace*{2em}{\color{red}\textbf{(Remaining 7 multi-profile evaluation queries synthesized by \scenegen{})}}\\[4pt]
directions: \\
\hspace*{1em}- START -> test\_model\\
\hspace*{1em}- test\_model -> END\\[4pt]
evaluation:\\
\hspace*{1em}name: judge\\
\hspace*{1em}model: judge\_model\\
\hspace*{1em}prompt: \\
\hspace*{2em}- role: user\\
\hspace*{3em}content: |\\
\hspace*{4em}{\color{blue}\textbf{You are an expert in education, specializing in evaluating scenarios for Chinese-Elementary-Ancient Poetry Enlightenment. Please rate the following response using a 1-5 scale...}}\\
\hspace*{1em}format:\\
\hspace*{2em}{\color{blue}\textbf{- field: Accuracy of Poetry Knowledge}}\\
\hspace*{2em}{\color{blue}\textbf{- field: Heuristic Teaching Guidance}}\\
\hspace*{2em}{\color{blue}\textbf{- field: Grade-level Language Adaptability}}\\
\hspace*{2em}{\color{blue}\textbf{- field: Completeness of Instructional Stages}}\\
\hspace*{2em}{\color{blue}\textbf{- field: Student Profile Alignment}}\\
\hspace*{2em}{\color{blue}\textbf{- field: Format Instruction Compliance}}\\
\hspace*{2em}{\color{blue}\textbf{- field: Personalized Problem Diagnosis}}\\
\hspace*{2em}{\color{blue}\textbf{- field: Stimulation and Guidance of Imagination}}\\
\hspace*{2em}{\color{blue}\textbf{- field: Integration of Traditional Cultural Values}}\\
\hspace*{2em}{\color{red}\textbf{(Comprehensive fine-grained rubrics generated by \scenegen{})}}
\end{minipage}}
\caption{YAML configuration samples for the elementary school ancient poetry enlightenment scenario.}
\label{fig:yaml-case-study}
\end{figure*}

\end{document}